\documentclass[runningheads]{llncs}
\usepackage{graphicx}

\usepackage{tikz}
\usepackage{comment}
\usepackage{amsmath,amssymb} 
\usepackage{color}
\usepackage{orcidlink}

\usepackage[accsupp]{axessibility}  

\usepackage{xcolor}
\definecolor{patrick_color}{rgb}{.6,.4,.05}
\definecolor{chengcheng_color}{rgb}{.5,.7,.1}
\definecolor{chris_color}{rgb}{0,0.35,0}
\definecolor{chengde_color}{rgb}{0,0,1}
\definecolor{charlie_color}{rgb}{0,0,0.8}
\definecolor{samarth_color}{rgb}{0.75,0.25,0.0}
\definecolor{new_color}{rgb}{0.8,0.0,0.0}

\newcommand{\etal}{\textit{et al}.}

\usepackage{adjustbox}
\usepackage{array}

\newcolumntype{R}[2]{%
    >{\adjustbox{angle=#1,lap=\width-(#2)}\bgroup}%
    l%
    <{\egroup}%
}

\usepackage{xcolor,pifont}
\newcommand*\colorcheck[1]{%
  \expandafter\newcommand\csname #1check\endcsname{\textcolor{#1}{\ding{52}}}%
}
\colorcheck{blue}
\colorcheck{green}
\colorcheck{red}

\newcommand*\colorx[1]{%
  \expandafter\newcommand\csname #1x\endcsname{\textcolor{#1}{\ding{55}}}%
}
\colorx{blue}
\colorx{green}
\colorx{red}

\begin{document}
\pagestyle{headings}
\mainmatter
\def\ECCVSubNumber{6515}  

\title{PressureVision: Estimating Hand Pressure from a Single RGB Image} 


\titlerunning{PressureVision: Estimating Hand Pressure from a Single RGB Image}
%
\author{
Patrick Grady\inst{1}\orcidlink{0000-0002-7248-8178}
\and Chengcheng Tang\inst{2}\orcidlink{0000-0002-4875-6670}
\and Samarth Brahmbhatt\inst{3}\orcidlink{0000-0002-3732-8865}
\and Christopher D. Twigg\inst{2}\orcidlink{0000-0003-3778-2520}\index{Twigg, Christopher D.}
\and Chengde Wan\inst{2}\orcidlink{0000-0003-1762-7849}
\and James Hays\inst{1}\orcidlink{0000-0001-7016-4252}
\and Charles C. Kemp\inst{1}\orcidlink{0000-0003-4720-1136}\index{Kemp, Charles C.}
}
\authorrunning{P. Grady et al.}
%
\institute{
Georgia Institute of Technology\\
\and Meta Reality Labs\\
\and Intel Labs
}

\maketitle

\setcounter{footnote}{0}

\begin{abstract}
People often interact with their surroundings by applying pressure with their hands. While hand pressure can be measured by placing pressure sensors between the hand and the environment, doing so can alter contact mechanics, interfere with human tactile perception, require costly sensors, and scale poorly to large environments. We explore the possibility of using a conventional RGB camera to infer hand pressure, enabling machine perception of hand pressure from uninstrumented hands and surfaces. The central insight is that the application of pressure by a hand results in informative appearance changes. Hands share biomechanical properties that result in similar observable phenomena, such as soft-tissue deformation, blood distribution, hand pose, and cast shadows. We collected videos of 36 participants with diverse skin tone applying pressure to an instrumented planar surface. We then trained a deep model (PressureVisionNet) to infer a pressure image from a single RGB image. Our model infers pressure for participants outside of the training data and outperforms baselines. We also show that the output of our model depends on the appearance of the hand and cast shadows near contact regions. Overall, our results suggest the appearance of a previously unobserved human hand can be used to accurately infer applied pressure. Data, code, and models are available online\footnote{\url{https://github.com/facebookresearch/pressurevision}}.
\end{abstract}

\section{Introduction}


Humans often interact with their surroundings by applying pressure with their hands. Given the importance of hand pressure, methods that enable the machine perception of this quantity could have broad applications. Traditionally, measuring the pressure a hand exerts has been accomplished with physical sensors that sit between the hand and contact surface. This includes sensors worn on the hand, such as pressure-sensitive gloves, and sensors mounted to the environment, such as arrays of pressure sensors.

\begin{figure}
\begin{center}
   \includegraphics[width=1.0\linewidth]{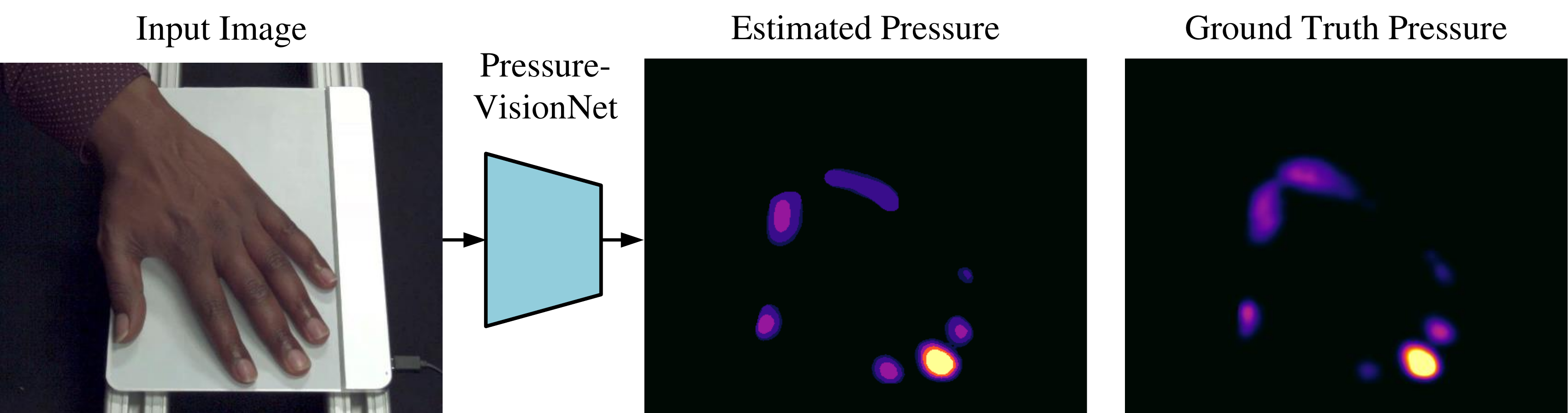}
\end{center}
\caption{PressureVisionNet takes an RGB image as input and outputs a pressure image with estimates of pressure applied by the hand to a planar surface. Each pixel of the pressure image is an estimate of hand pressure for the corresponding pixel in the RGB image. Black, purple, and yellow represent zero, low, and high pressure.}
\label{fig:full_pipeline}
\end{figure}

While physical sensors are accurate and robust, they have drawbacks. Sensors between the hand and the environment alter surface properties, changing appearance and interfering with contact mechanics relevant to human manipulation and tactile sensing. Sensors attached to hands can also be uncomfortable. Sensors mounted to environments require large numbers of sensing elements to cover modest areas, such tabletops, at high resolution and can be difficult to apply to varied surfaces.

Cameras have the potential to economically cover surfaces with virtual pressure sensors at high spatial resolution and do so without requiring the application of cumbersome instrumentation. Much like markerless pose estimation has enabled new applications by inferring human kinematics from RGB images, \textit{visual hand pressure estimation} has the potential to be widely applied. For example, modern augmented and virtual reality headsets have cameras that could be used to perceive hand pressure, turning a wall into a giant touchscreen, or a tabletop into piano keyboard. 

As a first step towards general visual hand pressure estimation, we investigate the feasibility of estimating hand pressure from RGB images under controlled conditions. To the best of our knowledge, our work is the first to demonstrate the feasibility of this approach. In particular, we address two critical questions: 1) Can appearance-based inference estimate hand pressure? 2) Can visual hand pressure inference generalize to previously unseen hands from a diverse population? We use a planar pressure sensing array to measure ground truth pressure during natural interactions from 36 participants with diverse skin tones. 

We developed PressureVisionNet to infer hand pressure from a single RGB image. PressureVisionNet performed well with unseen participants, providing evidence that the visual signals used for pressure estimation are shared across people. Our sensitivity analysis also indicates that PressureVisionNet relies on the appearance of the hand close to regions of contact, where we expect applied pressure to most influence hand appearance. Similarly, PressureVisionNet performs poorly at estimating pressure for visually occluded parts of the hand. We also provide a demonstration of PressureVisionNet inferring plausible pressure given conditions outside of the training data. Our demonstrations involve a smartphone camera observing an uninstrumented tabletop with ambient office lighting being contacted by people outside of the training set.


Our central insight is that hands display subtle visual cues that indicate the presence of pressure. As the hand applies pressure to a surface, the surface applies equal and opposite pressure to the hand. As such, the hand itself can serve as a physical probe that changes in predictable ways when pressure is applied to it, and the camera can observe this probe from a distance to infer the associated pressure. The shape of the hand can also result in informative shadows cast on the surface.

\begin{figure}[t]
\begin{center}
   \includegraphics[width=0.8\linewidth]{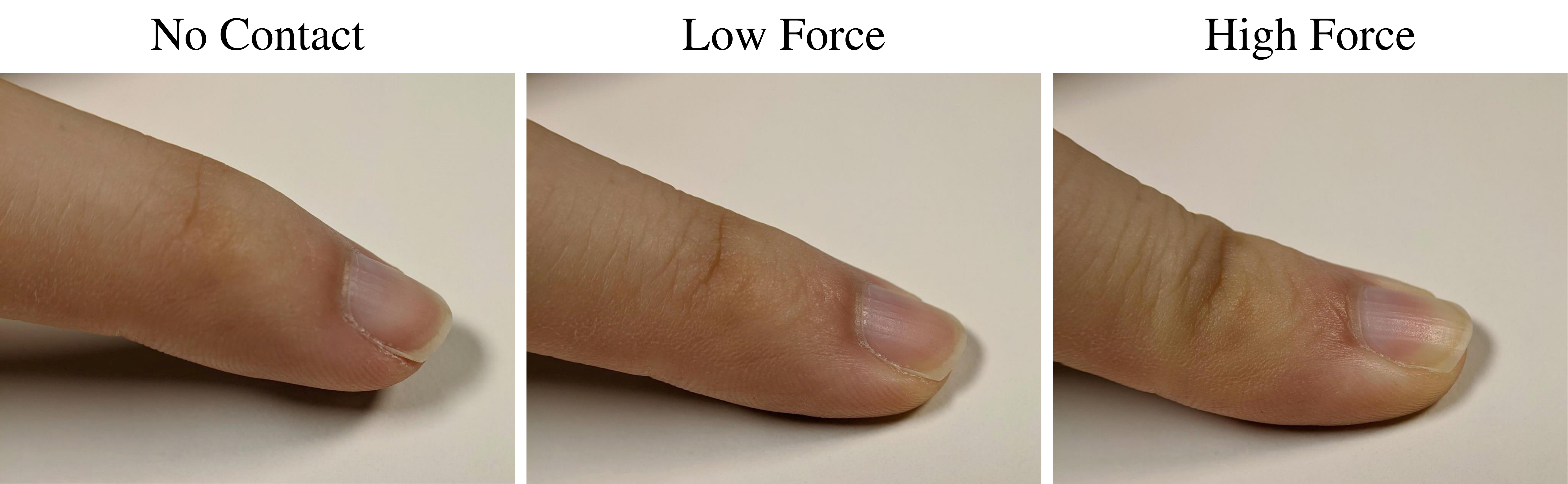}
\end{center}
\caption{Fingers display visible cues indicative of applied pressure. In the no-contact case, shadows are diffuse and the finger can be less sharp due to motion blur. As the finger makes contact, the skin at the fingertip loses color, the distal joint hyperextends, the texture of the skin changes, and the finger pad expands.}
\label{fig:finger_blowup}
\end{figure}

Hands are complex, non-rigid appendages with multiple types of soft tissues and fluids surrounding an interior skeleton that is often modeled as having 27 degrees of freedom. Pressure applied to the hand deforms tissues, moves fluids, and changes the configuration of the joints. The active application of pressure also involves contraction of hand muscles and tensile force applied by muscles in the forearm via tendons. These biomechanical processes result in changes to the appearance of the hand in visible light, including changes to surface geometry, color, and texture (Figure \ref{fig:finger_blowup}).


In summary, our paper makes the following contributions:
\begin{itemize}
    \item We propose a novel task, \textit{visual hand pressure estimation}.
    \item We present PressureVisionNet and show that it can infer hand pressure from a single RGB image and generalize to new people. 
    \item We release PressureVisionDB, a dataset of 36 participants with paired pressure and image data.
    \item We release our trained models and code.
\end{itemize}

\section{Related Work}

\subsubsection{Physical Sensors for Force Measurement}
Sensors to directly measure the force that a hand exerts generally fall into two categories: sensors on the hand and sensors on the surface being touched. Gloves instrumented with flexible force sensors can directly measure the force that the hand exerts \cite{buscher2015flexible,DBLP:journals/nature/SundaramKLZ0M19}. While sensors on the hand work for a wide range of objects, they can inhibit natural manipulation, reduce the person's tactile sensation, and often do not cover the entire hand. Commercially available systems are also expensive \cite{pps_tactileglove,tekscan_grip}.

Alternatively, objects can be instrumented to capture force information \cite{bhirangi2021reskin,brahmbhatt2020grasp,senselmorph,pham2017hand}. While these methods can accurately measure pressure, they change the properties of the surfaces they cover, can be challenging to mount on curved surfaces, and scale poorly to larger surfaces.


A variety of fingertip pressure sensors have been developed for robotic grippers. These internal sensors sometimes observe the deformation of a soft exterior \cite{ward2018tactip,yuan2017gelsight} or monitor an internal fluid \cite{wettels2008biomimetic}.


\subsubsection{Inferring Contact and Force from Vision}
One class of methods uses physics to infer the contact forces given the contact points and object trajectory. Methods proposed to determine the contact points include neural networks \cite{ehsani2020use,li2019motionandforce} or by combining markerless tracking of the hands with mesh-object intersection~\cite{pham2015forcesensing,pham2017hand,rogez2015understanding}.  These methods complement our own because they can infer contact that is occluded or out-of-view, but methods that rely on the accelerations of hand-held objects cannot generalize to interactions with relatively immobile surfaces like tables.

Additional work studies force application in general human-environment interactions such as sitting and standing~\cite{scott2020image,zhang2020_generatingpeople,zhu2016inferrring_forces}.
Clever \etal~use depth data to regress the amount of pressure between a human at rest and a mattress \cite{clever2021bodypressure}. They use a neural network to estimate the pose of a body on the bed in addition to pressure.

\subsubsection{Predicting Pressure using Hand and Surface Appearance} As the hand applies force, blood in the surrounding tissue is displaced. This effect is visible at the fingertip and underneath the fingernail, where a whitening of the tissue can be observed. Various techniques have been proposed to estimate fingertip force using optical sensors focused on this effect \cite{fingertip_color,mascaro2001photoplethysmograph,mascaro2004}.


The soft tissue in the palm and pads of the fingers deforms under applied force \cite{perez2013stiffness}. This deformation is accompanied by expansion in other areas, often visible as a widening of the pads of the fingers.  Hwang \etal~\cite{hwang2017inferring} use surface deformation to infer contact force, but support only unoccluded point contacts.
Johnson \etal~\cite{johnson2009retrographic} use the deformation of the hand to perform a high-resolution reconstruction of the object surface.  

Cast shadows provide important cues in human perception of depth in a 3D scene \cite{hu2002visual,hu2000visual,cast_shadows_psychology}. Researchers have used visual observations of shadows for closed-loop control of a robot \cite{cast_shadows}.

It has additionally been demonstrated that videos of faces can be used to identify if participants are squeezing an object tightly~\cite{asadi2020computer}. 

\subsubsection{Amplifying Imperceptible Visual Cues} 
Various techniques have been presented to extract subtle cues from images or video which are typically not noticed by the human eye.
Wu \etal~present Eulerian Video Magnification \cite{video_magnification}. They magnify periodic signals to make them clearly visible, including minute changes in skin color due to bloodflow. Davis \etal~\cite{visual_microphone} are able to amplify tiny subpixel vibrations in videos to reconstruct audio from visual data.

\subsubsection{Contact for Grasping and Pose Estimation}
Contact between hands and objects is important for grasping and manipulation. Estimating where contact should occur is often a first step for planning a robotic grasp~\cite{brahmbhatt2019contactgrasp,chu2018real,rosales2011global,saxena2008robotic}
or creating anthropomorphic animation~\cite{karunratanakul2020grasping,synthesisOfHandManipulationUsingContactSampling,zhang2021manipnet}.
For pose estimation during hand-object interaction~\cite{garcia2018first,hampali2020honnotate}, using contact to enforce consistency between hands and objects improves pose plausibility and accuracy~\cite{contactopt,hasson2020leveraging,hasson2019learning,romero2010hands,tzionas2016capturing}.
In a broader context, contact is used when reasoning about people interacting with the surrounding environment~\cite{clever2020bodies,holden2020learned,narasimhaswamy2020detecting,starke2020local,zhang2020_generatingpeople}.

\subsubsection{Contact for Human-Computer Interaction}
Detecting contact with ordinary surfaces can be useful for building human-computer interfaces.
MRTouch \cite{mrtouch} uses depth data to identify touches on flat surfaces for the Microsoft HoloLens. The method first fits a 3D plane to the wall or tabletop surface and thresholds the distance between the plane and the fingertip.
TapID \cite{meier2021tapid} uses subtle accelerations from a pair of IMU sensors on the wrist to detect contact between fingers and surfaces.
TouchAnywhere \cite{niikura2014anywhere} detects touch via heuristics by estimating the intersection between finger and its shadow, but only detects binary contact and does not include pressure ground truth.

\begin{table}
\caption{Datasets for contact between hands and surfaces: GRAB~\cite{GRAB:2020} infers dynamic contact via geometric models and marker-based motion capture. TactileGloves~\cite{DBLP:journals/nature/SundaramKLZ0M19} measures dynamic pressure with a sensorized glove. ContactDB~\cite{contactdb} and ContactPose~\cite{brahmbhatt2020contactpose} measure static contact via thermal imaging. Ours, PressureVisionDB, measures pressure with a sensorized plane and provides registered and synchronized RGB images.}
\centering
\begin{tabular}{c|c|c|c|c}
    &\textbf{\begin{tabular}{@{}c@{}}Diverse\\Surfaces\end{tabular}}&\textbf{\begin{tabular}{@{}c@{}}Bare\\Hands\end{tabular}} &\textbf{Images}&\textbf{Pressure}
    \\\hline
    GRAB~\cite{GRAB:2020} & \greencheck & \redx & \redx & \redx \\\hline
    TactileGloves~\cite{DBLP:journals/nature/SundaramKLZ0M19} & \greencheck & \redx & \redx & \greencheck \\\hline
    ContactDB~\cite{contactdb} & \greencheck & \greencheck & \redx & \redx \\\hline
    ContactPose~\cite{brahmbhatt2020contactpose} & \greencheck & \greencheck & \greencheck & \redx \\\hline
    PressureVisionDB (Ours) & \redx & \greencheck & \greencheck & \greencheck \\\hline
\end{tabular}
\label{tab:dataset_comparison}
\end{table}

\subsubsection{Hand-Object Contact Datasets}
GRAB~\cite{GRAB:2020} uses optical motion capture to capture hand and object pose and indirectly infers contact using mesh proximity.
ContactDB~\cite{contactdb} measures high-resolution contact on a variety of objects.  After participants grasp an object, the thermal imprint from the hand is measured from the object surface using a thermal camera. 
ContactPose~\cite{brahmbhatt2020contactpose} adds paired images, object poses, and hand poses to ContactDB. ContactDB and ContactPose do not provide pressure measurements and only provide contact for static grasps. 

As shown in Table~\ref{tab:dataset_comparison}, our dataset is unique compared with these datasets, providing RGB images paired with high-resolution, dynamic pressure images.

\section{The PressureVisionDB Dataset}

We collected PressureVisionDB, a novel dataset of bare hands interacting with a pressure-sensitive object. 
The dataset includes 16 hours of data collected from 36 participants with pressure data and RGB images from four synchronized cameras, totalling 64 hours of RGB video data.

\subsection{The Capture Setup}

\begin{figure*}
\begin{center}
   \includegraphics[width=1.0\linewidth]{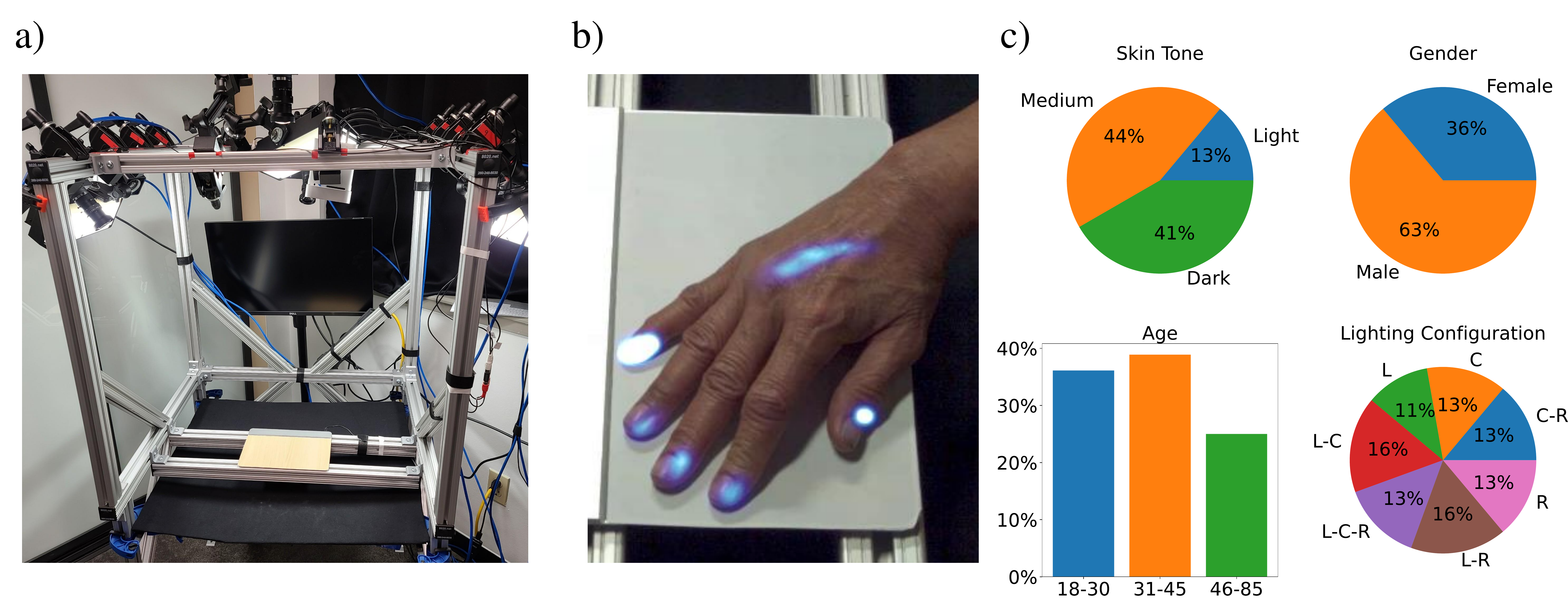}
\end{center}
\caption{a) The data capture rig used to collect PressureVisionDB. Participants reach into the front opening to make contact with the elevated pressure sensor. b) The sensorized plane records a pressure image, which is projected into image space for training and evaluation. Pressure is visualized as bright blue regions. c) PressureVisionDB includes participants with diverse skin tones, ages, and genders under varied lighting conditions.}
\label{fig:dataset_multifig}
\end{figure*}

We assemble a rigid frame to which we attached cameras, lights, and a pressure sensing surface (Figure \ref{fig:dataset_multifig}a). Participants reached through one side of the cube-shaped frame to place their hands on the planar sensing surface. 




\subsubsection{Pressure Sensor} To measure hand pressure, we used a Sensel Morph \cite{senselmorph} (Figure \ref{fig:dataset_multifig}b). The sensor is a planar surface featuring a grid of force-sensitive resistor (FSR) pixels. The sensor produces a 185x105 ``pressure image'', with each pixel having a pitch of 1.25mm. The sensor produces diffuse readings below 0.5 kPa, which we use as the minimum effective pressure, and the 99th percentile pressure recorded in the dataset is 82 kPa. 

We covered the sensor surface with a white vinyl overlay for most participants, and covered it with a wood-textured overlay for seven participants. 
We mounted the sensor above the table surface to allow participants to reach underneath the sensor for pinching and grasping actions (Figure \ref{fig:dataset_multifig}a). PressureVisionDB only measures pressure across this single planar surface, and we leave the capture of hand pressure on more diverse shapes to future work. Additionally, the sensor measures pressure normal to the plane, so we only consider this component of pressure.



\subsubsection{Cameras} 
The capture setup uses four synchronized and calibrated OptiTrack Prime Color cameras to capture 1080p RGB frames. We mounted the cameras to provide overhead views of the scene along with three OptiTrack eStrobe light sources. Each light can be turned on or off to achieve 8 different lighting conditions. The lighting condition was changed after each participant. 


\subsubsection{Participants} 
\begin{figure*}
\begin{center}
   \includegraphics[width=0.95\linewidth]{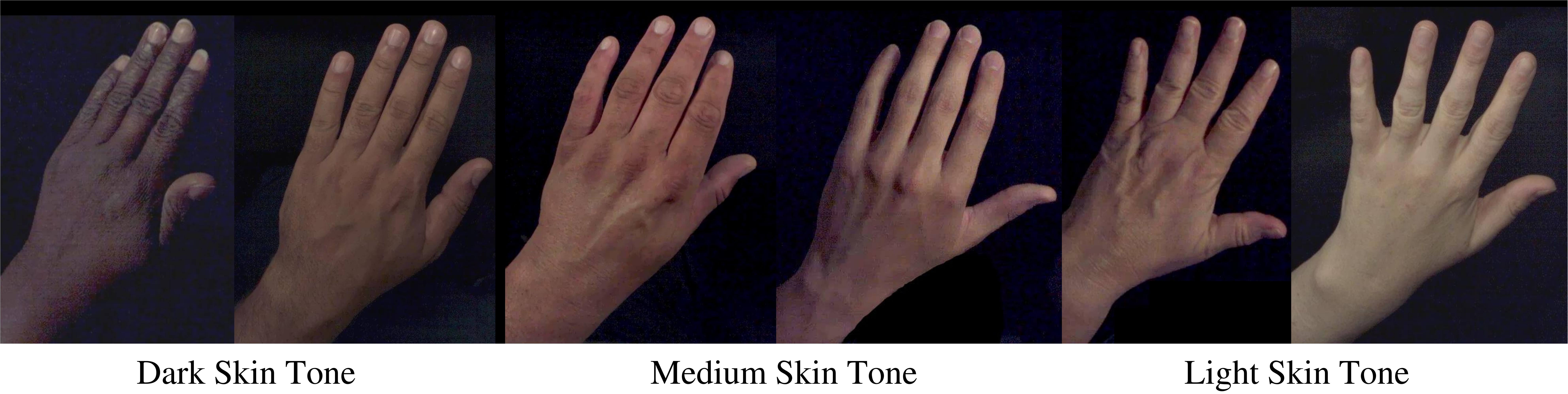}
\end{center}
\caption{PressureVisionDB includes data from participants with diverse skin tones.}
\label{fig:sample_hands}
\end{figure*}

PressureVisionDB was collected from a diverse set of adults (Figure \ref{fig:sample_hands}) with a range of skin tones, ages, and genders (Figure \ref{fig:dataset_multifig}c). Participants self-reported their age and gender, and their skin tone were measured objectively. A Pantone RM200 colorimeter was used to classify the participant's skin tone in the Pantone SkinTone Guide \cite{pantone}. These skin tones were then divided into three categories: light, medium, and dark. 

\subsubsection{Protocol}
We developed a protocol to capture a variety of hand-surface interactions. Participants were asked to perform 36 different actions with one hand and then the other. Prior to each action, participants were shown a text description and an image of the action. The list of actions includes pressing with a single finger (index, thumb, or other), pressing with the whole hand, applying tangential force, grasping the edge of the sensor, and drawing with a finger. For many actions, participants were prompted to use one of three force levels: high force, low force, and no contact. For the no contact condition, participants moved their hand very close to the sensor, as if they intended to touch the sensor, but did not make contact.




\subsubsection{Ethics} 
Participants gave informed consent and were compensated for their time. The data was captured by a third-party corporation that specializes in data collection with human participants. The capture rig was designed so that no images of the participants' torsos or faces were collected. The third-party did not provide personally identifiable information to the research team.



At the inception of our project, we were concerned about the potential for darker skin tones to reduce performance \cite{cook2019demographic,krishnapriya2020issues,cavazos2020accuracy}. Melanin and hemoglobin both influence skin color in visible light and melanin concentration varies widely across people \cite{zonios2001skin}. As such, we prioritized recruiting participants with a wide range of skin tones  (Figure \ref{fig:dataset_multifig}c).

%


\section{Estimating Pressure and Contact from RGB}

We designed a deep network to infer hand contact and pressure. From an input RGB image, the network estimates a pressure image of the same size. The pressure image provides a pressure estimate for each pixel of the RGB image.






\subsection{Network Architecture}
\label{sec:network_arch}

We designed PressureVisionNet to estimate the location and magnitude of pressure between a hand and surface. The network is an encoder-decoder architecture which inputs a single RGB image, $I$, and outputs a pressure map, $\hat{P}=f(I)$. During training and evaluation, the ground truth pressure data from the sensor is projected into image space (Figure \ref{fig:dataset_multifig}b). Consequently, the network estimates hand pressure for each pixel of the input image, which results in an output pressure image that is the same size as the input image. 


Pressure estimation is treated as a classification problem. The pressure range is divided into nine bins placed in logarithmic space \cite{dehaene2003neural}. The network infers the pressure bin for each pixel of the output pressure image.  
We experimented with direct regression of pressure scalars, but found that this was outperformed by the classification approach.

We cropped images from each camera to include a margin around the pressure sensor and resized the images to 480x384 pixels. PressureNet uses a SE-ResNeXt50 \cite{resnet,squeeze-excitation,resnext} encoder, with weights from pretraining on ImageNet \cite{deng2009imagenet}. A feature pyramid network (FPN) \cite{fpn,segmentation_models_pytorch} decoder produces the output image. A cross-entropy loss was used during training. During inference, the network runs at 53 FPS using an RTX 3090 GPU. 

\begin{figure*}
\begin{center}
   \includegraphics[width=1.0\linewidth]{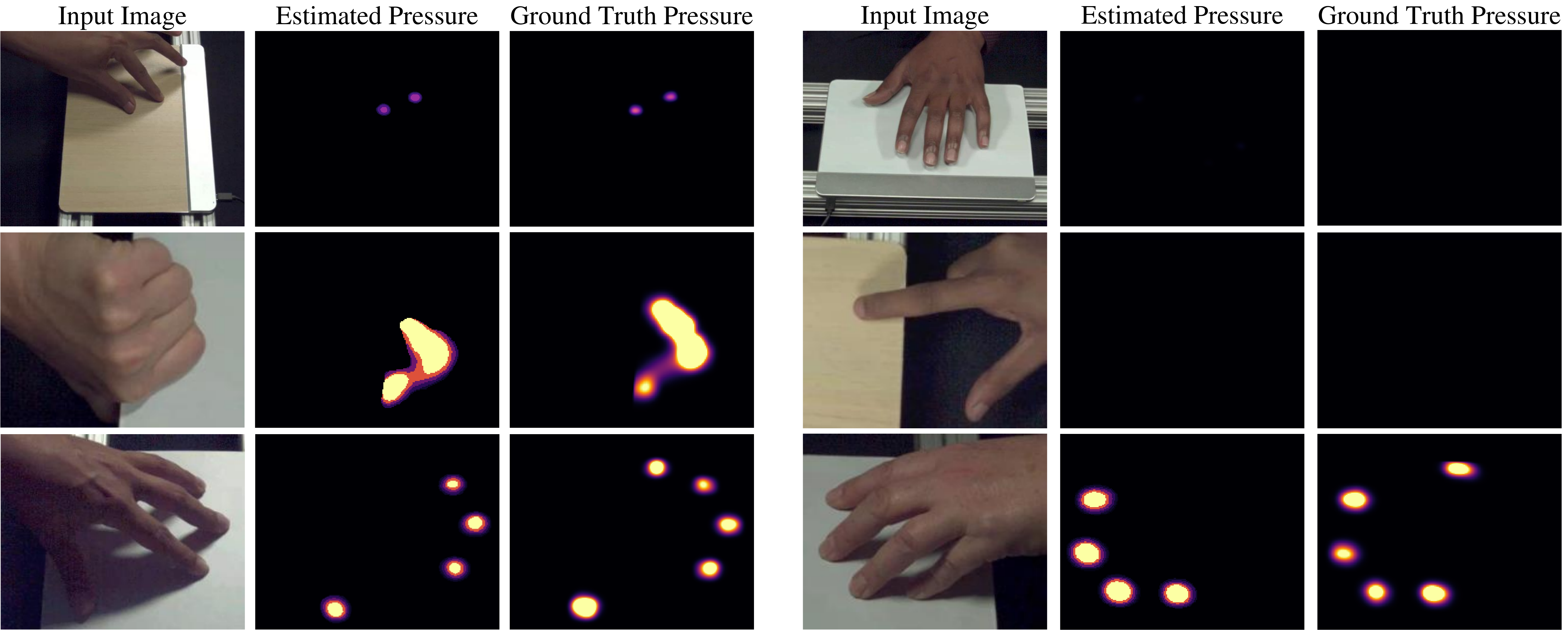}
\end{center}
\caption{Results from PressureVisionNet on held-out participants. Some images have been magnified for clarity. PressureVisionNet can accurately infer the location and magnitude of pressure from an RGB image. PressureVisionNet correctly identifies the lack of contact in ``no-contact'' actions (right, top two). The bottom row illustrates a common failure mode: no pressure is estimated for parts of the hand that are occluded.} 
\label{fig:main_results}
\end{figure*}

\subsection{Evaluation Metrics}

We considered four types of contact and pressure metrics. Contact is a binary quantity that describes if the hand and sensor are touching, while pressure is a scalar describing how hard the hand and sensor are pressing against each other. Contact maps $C$ are generated by thresholding the pressure map $P$ at a low value, $P_{th}=1.0$ kPa. Values greater than this are marked as \textit{in contact}.


\begin{itemize}
\item\textbf{Temporal Accuracy} To evaluate the \textit{temporal} accuracy with which the onset and termination of contact are estimated, if \textit{any} contact is present in the estimated and ground truth contact maps, $\hat{C}$ and $C$, the frame is marked as in contact. A frame is marked correct if the presence of contact is consistent in estimated and ground truth frames.

\item\textbf{Contact IoU} To evaluate the \textit{spatial} and \textit{temporal} accuracy of estimated contact, we computed the intersection over union (IoU) between the the binary contact images. This metric does not consider the magnitude of the estimated pressure, and is an upper bound on Volumetric IoU.

\item\textbf{Volumetric IoU} We propose the Volumetric IoU (Figure \ref{fig:volumetric_iou}), a novel metric that extends Contact IoU to evaluate the \textit{magnitudes} of pressure estimates in addition to their \textit{spatial} and  \textit{temporal} accuracy. Each 2D pressure image is converted into a 3D ``pressure volume", where the height of the volume is equal to the amount of pressure at that pixel. The Volumetric IoU can be calculated as:
\begin{align}
    IoU_{vol}=\frac{\sum^{i,j}min(P_{i,j}, \hat{P}_{i,j})}{\sum^{i,j}max(P_{i,j}, \hat{P}_{i,j})}
\end{align}

\begin{figure}
\begin{center}
   \includegraphics[width=0.53\linewidth]{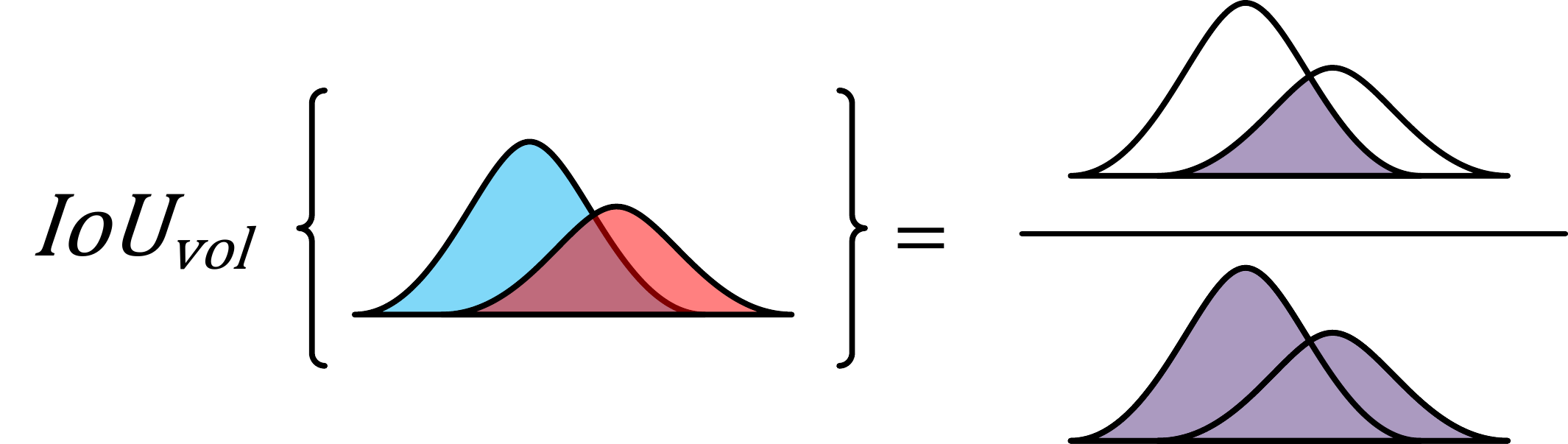}
\end{center}
\caption{Volumetric IoU quantifies the agreement between two pressure images by treating them as pressure volumes.}
\label{fig:volumetric_iou}
\end{figure}



\item\textbf{Mean Absolute Error} To evaluate the accuracy of estimated pressure in \textit{physical units}, we calculate mean absolute error (MAE) over each pixel. As most of the dataset pressure images consist of zeros, these numbers are close to zero.
\end{itemize}




\subsection{Dataset Splits}
\label{sec:cross_val}


The dataset consists of 36 participants. Prior to using the dataset to develop PressureVisionNet, we selected 6 participants spanning skin tone and demographics for a held-out test set. We used data from the remaining 30 participants for training and validation.




\section{Results}

Our primary goal was to investigate the potential to infer hand pressure from a single RGB image. In addition to characterizing overall performance, we focused on three questions. First, we considered how well performance can generalize to people outside of the training set, since feasibility depends on the existence of shared properties across hands and human behavior. Second, we considered whether the appearance of the hand and its cast shadows were used for inference. Third, we considered whether a trained model can produce reasonable estimates for images captured in an unseen environment.



\begin{figure}
\begin{center}
   \includegraphics[width=0.8\linewidth]{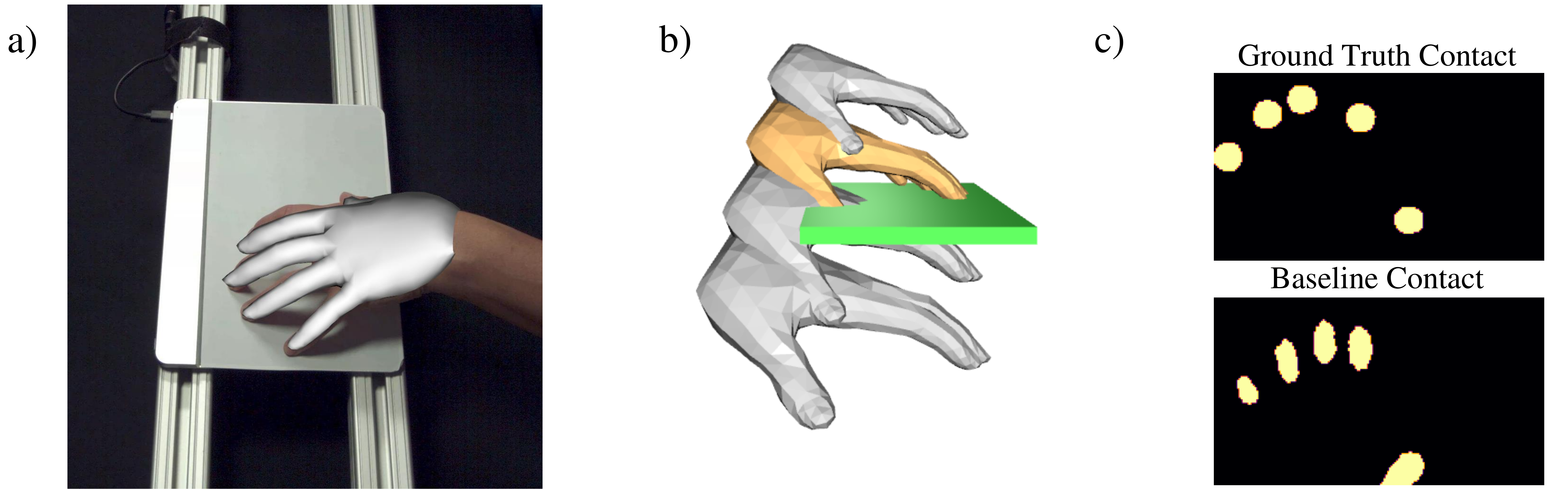}
\end{center}
\caption{a) A 3D pose estimator is used to estimate hand pose. b) As monocular pose estimators have difficulty in estimating true hand scale, hand scale is swept through a range of values (selected scales are visualized, optimal scale is colored). c) Ground truth contact is compared to contact calculated by mesh intersection.}
\label{fig:baseline_fig}
\end{figure}

\subsection{Baseline Models}

Since inferring hand pressure from an RGB image is a new task, there were no existing methods available for direct comparison. To provide context for our numeric results, we created a baseline model for comparison. 

\label{sec:pose_estimator_baseline}
3D models of the human body have been used to infer contact and pressure in other contexts \cite{GRAB:2020,contactopt,clever2021bodypressure}. Our use of a single RGB image makes this approach more challenging. Prior work used multiple cameras, depth cameras, and motion capture systems to obtain high fidelity 3D models. We used FrankMocap \cite{rong2021frankmocap} to produce a 3D pose estimate with a free parameter for the scale of the hand. As the true hand scale is not known, for each sequence we swept the hand scale with a discretization that corresponds to sub-millimeter adjustment in depth, and find the scale that maximizes the Contact IoU with respect to ground truth (Figure \ref{fig:baseline_fig}). Contact was estimated by finding intersection between the hand and sensor meshes.

\subsection{Can inference succeed with new people?}

\begin{table*}
\caption{PressureVisionNet outperforms the 3D Pose Baseline on a test set of unseen participants.} 
\centering
\begin{tabular}{c|c|c|c|c}
    \textbf{Method} & \textbf{Temporal Acc} & \textbf{Contact IoU} & \textbf{Vol. IoU} & \textbf{MAE}\\\hline
    Zero Guesser & 53.7\% & 0.0\% & 0.0\% & 51.9 Pa \\\hline
    3D Pose Baseline \cite{rong2021frankmocap} & 78.1\% & 13.0\% & - & -\\\hline
    \textbf{PressureVisionNet} & 96.2\% & 55.8\% & 41.3\% & 39.9 Pa \\\hline
\end{tabular}
\label{tab:overall_results}
\end{table*}

Table \ref{tab:overall_results} shows performance on the $\approx$490k images in our test set, which includes frames from all four cameras, five distinct lighting conditions, and 6 participants performing 36 actions. None of the participants in our test set were in our training set, so performance indicates how well our approach generalizes to new people.

\begin{table}
\caption{Participants were prompted to perform actions with various force levels. Estimated forces from PressureVisionNet correlate with ground truth (GT).}
\centering
\begin{tabular}{c|c|c}
    \textbf{Force Requested} & \textbf{Mean GT} & \textbf{Mean Est.} \\
    \textbf{to Participant} & \textbf{Force} & \textbf{Force} \\\hline
    High Force & 8.16 N & 5.73 N  \\\hline
    Low Force & 3.24 N & 3.63 N  \\\hline
    No Contact & 0.00 N & 0.04 N  \\\hline
\end{tabular}
\label{tab:adversarial_touch}
\end{table}

PressureVisionNet outperformed our baseline model. Its discretized representation for pressure limits its best possible Volumetric IoU to $81\%$ and MAE to $10.9$ Pa. The \textit{Zero Guesser} always outputs a zero-pressure image. It achieved $53.7\%$ temporal accuracy since no pressure was recorded for the majority of frames in the dataset.
The \textit{3D Pose Estimator} had generally low performance.
PressureVisionNet performed well with non-contact images and inferred higher pressures on average when participants were instructed to apply more force (Table \ref{tab:adversarial_touch} and Figure \ref{fig:main_results}).


 



\subsubsection{Skin Tone} 
\label{sec:skin_tone}
Skin tone is a significant source of variability across people. Due to the limited number of participants in each skin tone category, we used cross-validation. The training set is split into five folds, and the model is trained with one rotating fold left out for testing. As shown Table \ref{tab:skin_tone_results}, the highest performance was with dark skin tones.
We performed a two-sample Kolmogorov–Smirnov test between each pair of skin tone categories, and did not find a statistically significant difference.

\begin{table}
\caption{Performance across skin tones is found via cross validation. We did not find a statistically significant difference between categories.}
\centering
\begin{tabular}{c|c}
    \textbf{Skin Tone} & \textbf{Vol. IoU}\\\hline
    Light & 38.5\% \\\hline
    Medium & 37.2\% \\\hline
    Dark & 39.0\% \\\hline
\end{tabular}
\label{tab:skin_tone_results}
\end{table}



\subsection{Is hand-related appearance used for inference?}




We expected the appearance of the hand and cast shadows to provide visual cues for pressure inference. Our results suggest that this is the case. In addition to providing evidence for the underlying information used for inference, these results reduce the likelihood that PressureVisionNet is exploiting information specific to our data that would not be generalizable, such as unintended sensor artifacts.



\subsubsection{Sensitivity Analysis} We conducted a sensitivity analysis to identify spatial regions that strongly influence PressureVisionNet's output. Our method is similar to the Occlusion Sensitivity method from Zieler and Fergus \cite{occlusion_network_explanation}. We divide the input RGB image into a 48x48 grid of cells. For each cell $i,j$, we create a new image $\mathbf{B}_{i,j}(I)$ replacing the cell contents with its average color. We provide this image as an input to the network and calculate how much the output pressure changes. We create a sensitivity image by normalizing the following image, $S$, to be between 0 and 1.
\begin{align}
    S_{i,j} = ||f(\mathbf{B}_{i,j}(I)) - f(I)||_2
\end{align}

\begin{figure*}
\begin{center}
   \includegraphics[width=1.0\linewidth]{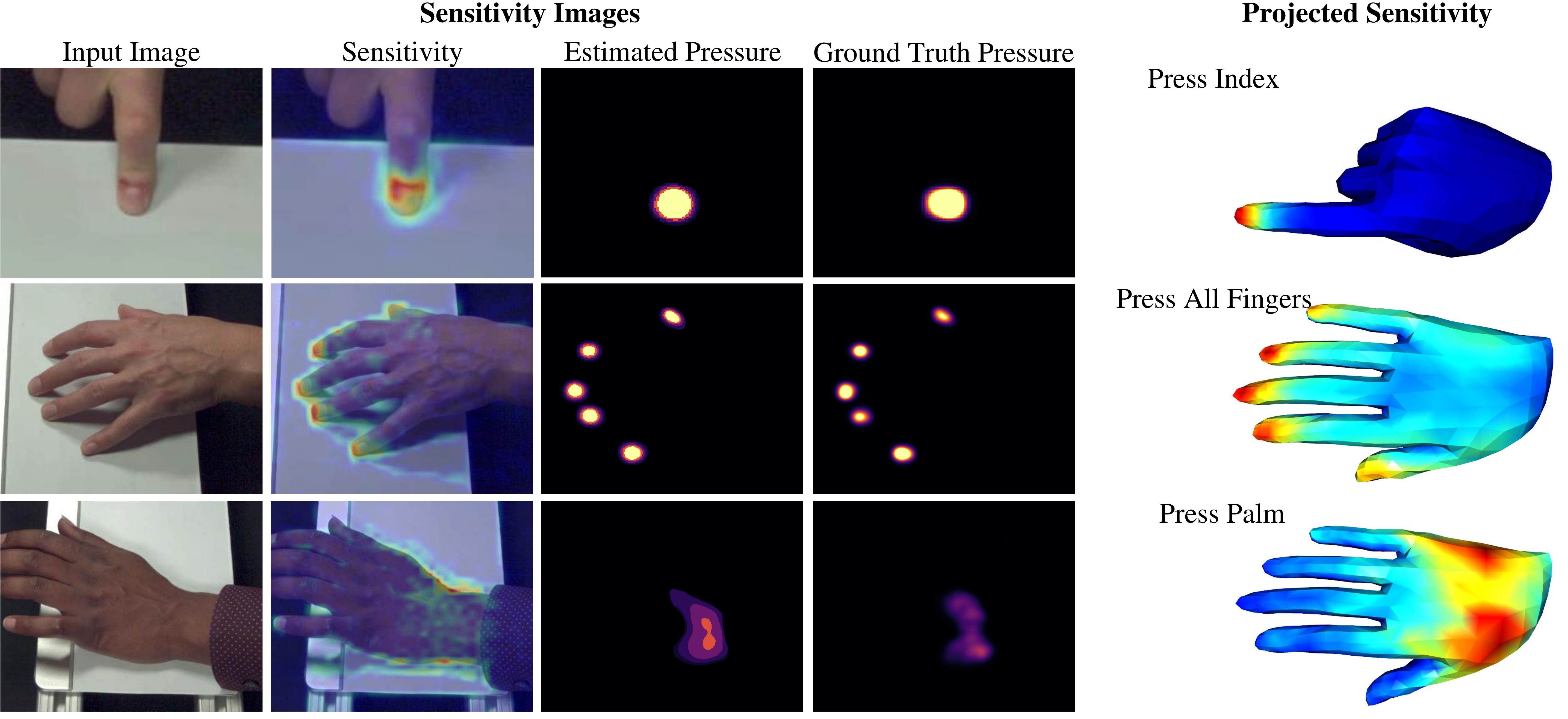}
\end{center}
\caption{Left: PressureVisionNet is sensitive to the appearance of the hand and shadows near regions of contact. Right: Results of projecting sensitivity images onto hand meshes and averaging over many frames for selected actions.}
\label{fig:saliency}
\end{figure*}

Figure \ref{fig:saliency} shows examples of resulting sensitivity images that indicate the model uses the appearance of the hand and cast shadows near regions of contact. To show this objectively, we also projected the sensitivity images from multiple frames onto the hand meshes generated for the 3D Pose Baseline. Figure \ref{fig:saliency} shows the results for three different actions that confirm that PressureVisionNet is highly sensitive to blurring of the hand near regions of contact. 



We also found a common failure mode that corroborates this result. As shown in the bottom row of Figure \ref{fig:main_results}, PressureVisionNet typically guesses zero pressure for parts of the hand that are occluded from view.

\subsubsection{Dependence on Actions} We found that PressureVisionNet's performance depended on the participant's action. For example, performance was highest with a single index finger action and lower with actions that apply pressure with the palm (Table \ref{tab:palm_vs_fingers_results}). Two relevant factors are likely the visibility of the part of the hand near the region of contact and the tendency for larger areas of ground truth pressure to result in larger errors. 

\begin{table}
\caption{The performance of PressureVisionNet on selected actions. Actions with lower visibility and larger contact areas tend to result in lower performance.}
\centering
\begin{tabular}{l|c}
    \textbf{Action name} & \textbf{Vol. IoU}\\\hline
    Press index, pull towards & 59.4\% \\\hline
    Press all fingers & 42.1\% \\\hline
    Press all fingers and palm & 35.4\% \\\hline
    Press palm & 33.2\% \\\hline
\end{tabular}
\label{tab:palm_vs_fingers_results}
\end{table}

\begin{figure}
\begin{center}
   \includegraphics[width=0.8\linewidth]{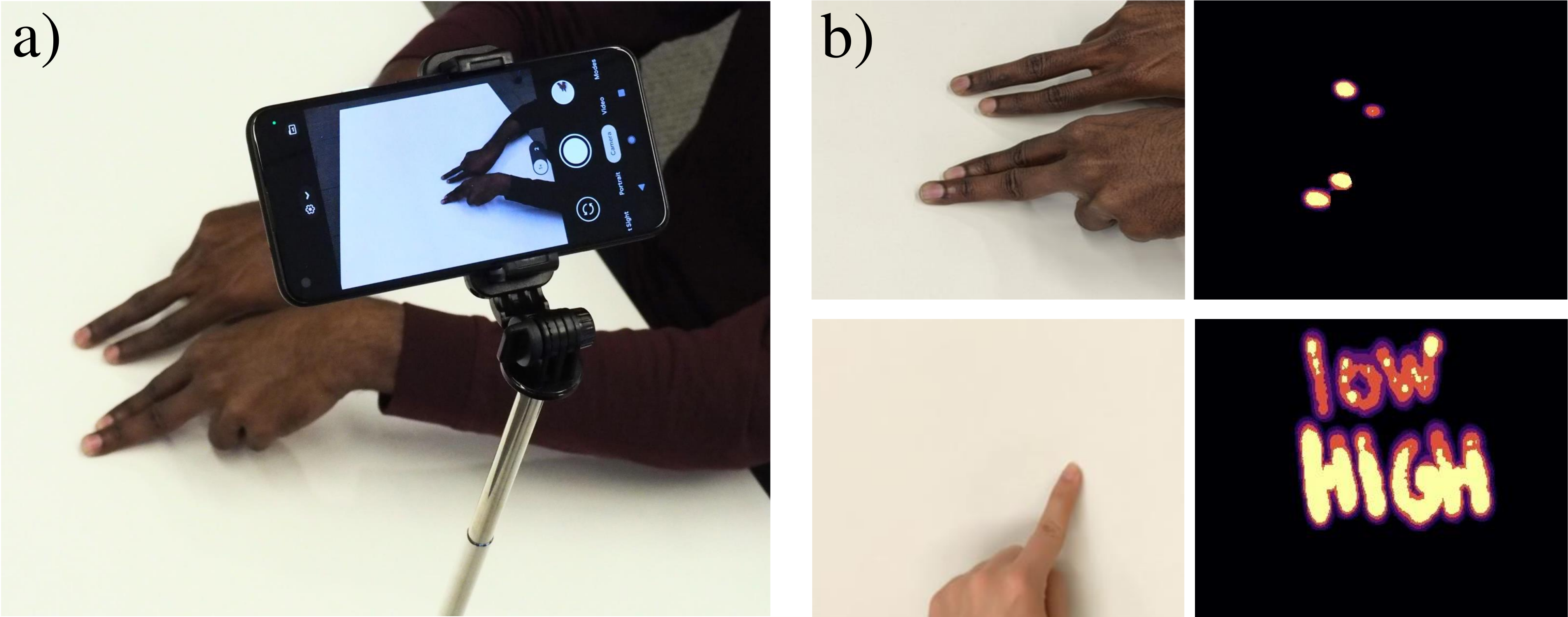}
\end{center}
\caption{PressureVisionNet may generalize to new environments. Results shown in b) were captured with a smartphone camera in unaltered office lighting on a tabletop. Pressure estimates can be accumulated over time to allow writing with a finger.}
\label{fig:touch_screen}
\end{figure}

\subsection{Is inference reasonable with new conditions?}
\label{sec:generalization}
We considered whether PressureVisionNet can produce reasonable estimates when given new RGB images acquired with conditions that differ from PressureVisionDB. For this question, we provide preliminary evidence based on images of hands not included in the dataset. We captured additional data with a smartphone camera in unaltered office lighting. The smartphone camera had different focal length, resolution, and color characteristics than the dataset cameras. The smartphone was mounted with a tripod to observe a normal white tabletop.

Figure \ref{fig:touch_screen} demonstrates that PressureVisionNet has some ability to generalize to new environments. The system can generalize to images with multiple hands, and can run over multiple frames in a video while accumulating inferred pressure estimates. In the bottom image, an author used the system to write with their index finger first with low pressure, then with high pressure. This illustrates a potential application and provides evidence that PressureVisionNet is not overly sensitive to the camera, the illumination, or the contact surface. 

\section{Conclusion}

We collected a novel dataset, PressureVisionDB, and developed a deep model, PressureVisionNet, to infer hand pressure from a single RGB image. Using this model, we provided evidence that hand pressure can be accurately inferred from a single RGB image of a hand from a previously unobserved person.
Our results suggest that the appearance of regions of the hand near regions of contact are especially informative when inferring pressure.






\subsubsection{Acknowledgements}
We thank Kevin Harris, Steve Miller, and Steve Olsen for their help in data collection, and Robert Wang, Minh Vo, Tomas Hodan, Amy Zhao, Kenrick Kin, Mark Richardson, and Cem Keskin for their advice.

\clearpage
%
%
\bibliographystyle{splncs04}
\bibliography{egbib}

\clearpage

\title{Supplementary - PressureVision: Estimating Hand Pressure from a Single RGB Image} 


\titlerunning{PressureVision: Estimating Hand Pressure from a Single RGB Image}
%
\author{
Patrick Grady\inst{1}\orcidlink{0000-0002-7248-8178}
\and Chengcheng Tang\inst{2}\orcidlink{0000-0002-4875-6670}
\and Samarth Brahmbhatt\inst{3}\orcidlink{0000-0002-3732-8865}
\and Christopher D. Twigg\inst{2}\orcidlink{0000-0003-3778-2520}\index{Twigg, Christopher D.}
\and Chengde Wan\inst{2}\orcidlink{0000-0003-1762-7849}
\and James Hays\inst{1}\orcidlink{0000-0001-7016-4252}
\and Charles C. Kemp\inst{1}\orcidlink{0000-0003-4720-1136}\index{Kemp, Charles C.}
}
\authorrunning{P. Grady et al.}
%
\institute{
Georgia Institute of Technology\\
\and Meta Reality Labs\\
\and Intel Labs
}

\maketitle

\section{Introduction}

This document provides additional details and qualitative results to supplement the main paper. Section \ref{sec:supp_db} provides more details about the PressureVisionDB dataset. Section \ref{sec:supp_network} gives more details about training PressureVisionNet, and Section \ref{sec:supp_images} provides further results.

\section{PressureVisionDB Details}
\label{sec:supp_db}

\subsection{Table of Actions}

During data collection of PressureVisionDB, participants complete a list of actions once for one hand, then repeat for the other hand. These actions are listed in Table \ref{tab:actions}.

\subsection{Data Sampling Rate}

Data from four OptiTrack Prime Color cameras is captured at 1080p resolution at 60 Hz, and 185x105 pressure images are collected from the Sensel Morph at approximately 115 Hz. Due to the large size of the dataset, the data is subsampled to 15 Hz for all experiments.

\begin{figure}
\begin{center}
  \includegraphics[width=0.6\linewidth]{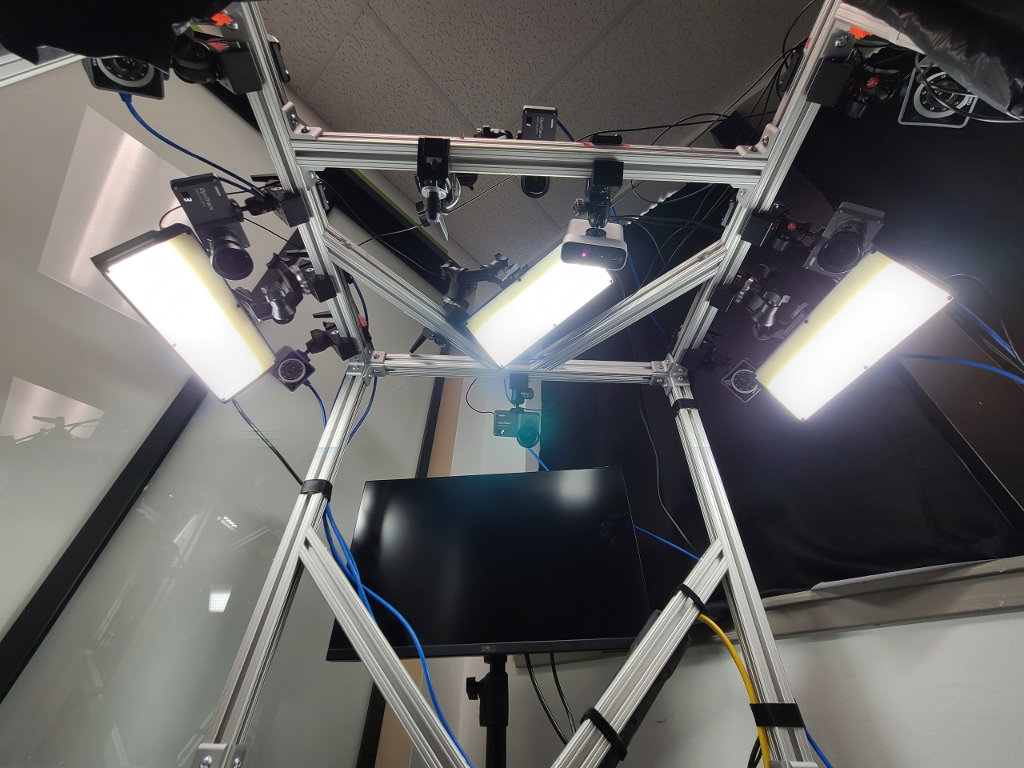}
\end{center}
\vspace{-8pt}
\caption{The top of the capture setup includes four OptiTrack Prime Color cameras and three OptiTrack eStrobe light sources. The light sources can be turned on or off in any combination.}
\label{fig:cage_top}
\end{figure}

\subsection{Lighting Combinations}

\begin{figure*}
\begin{center}
    \includegraphics[width=1.0\linewidth]{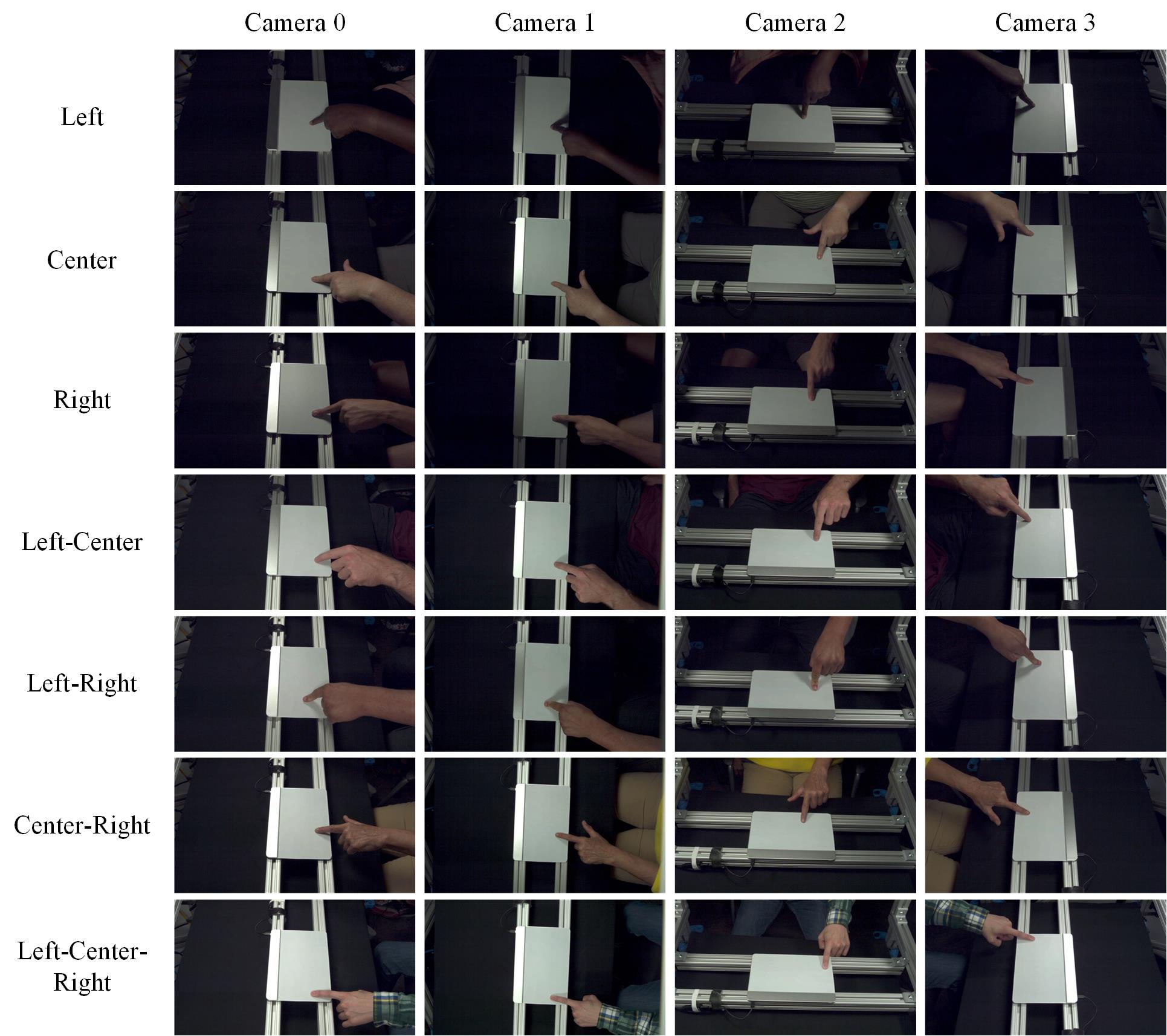}
\end{center}
\vspace{-8pt}
\caption{Data is captured with four cameras. Three lights are turned on and off to create seven lighting conditions.}
\label{fig:per_camera}
\end{figure*}

The capture setup used during the collection of PressureVisionDB includes four RGB cameras and three light sources: ``left'', ``center'', and ``right'' (see Figure \ref{fig:cage_top}). Figure \ref{fig:per_camera} shows images captured from each camera for each lighting condition.

\section{PressureVisionNet Details}
\label{sec:supp_network}

PressureVisionNet was implemented using a SE-ResNeXt50 encoder \cite{resnet,squeeze-excitation,resnext} and FPN decoder \cite{fpn} implementations from the \textit{segmentation-models-pytorch} project \cite{segmentation_models_pytorch}. The encoder network weights are initialized from pretraining on ImageNet~\cite{deng2009imagenet}, while the decoder weights are randomly initialized.

PressureVisionNet is optimized with the Adam optimizer with a batch size of 8. The network is trained with a learning rate of $1e-3$ for 100k iterations, then with a learning rate of $1e-4$ for 500k iterations. No data augmentation is performed during training.

\subsection{Input Images}

Images are cropped to include a 50-pixel border around the pressure sensor, and are resized to 480x384 pixels for the network. During experiments with reduced image quality (Section \ref{sec:supp_degraded}), images are first downsampled, then upsampled, such that the network architecture may remain constant.

The pressure image from the pressure sensor is warped into image space using a homography transform.

\section{Further PressureVisionNet Results}
\label{sec:supp_images}

\subsection{Qualitative Results}
\begin{figure*}
\begin{center}
  \includegraphics[width=1.0\linewidth]{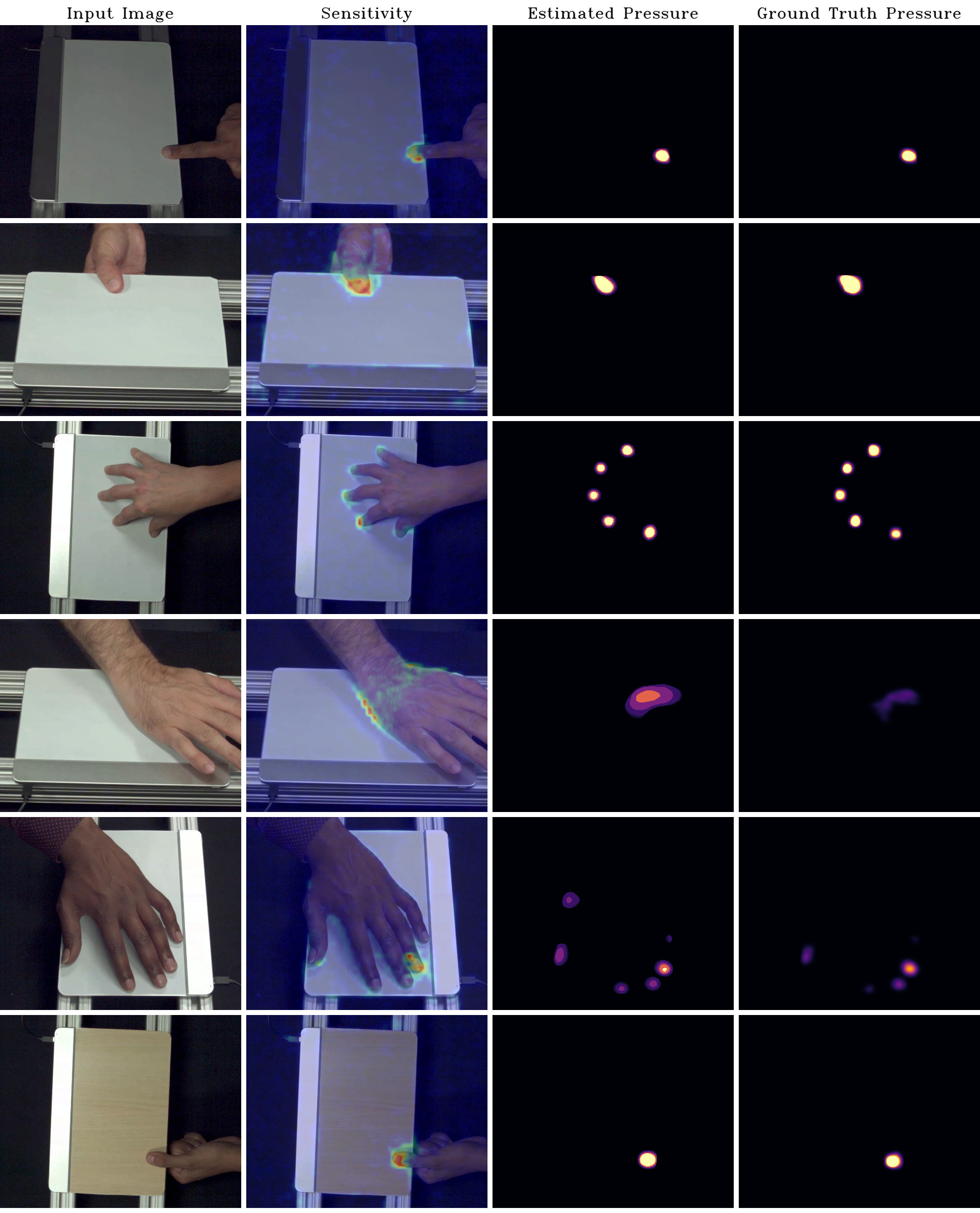}
\end{center}
\vspace{-8pt}
\caption{Results of PressureVisionNet selected \textit{randomly} from the test set.}
\label{fig:collage_0}
\end{figure*}

\begin{figure*}
\begin{center}
  \includegraphics[width=1.0\linewidth]{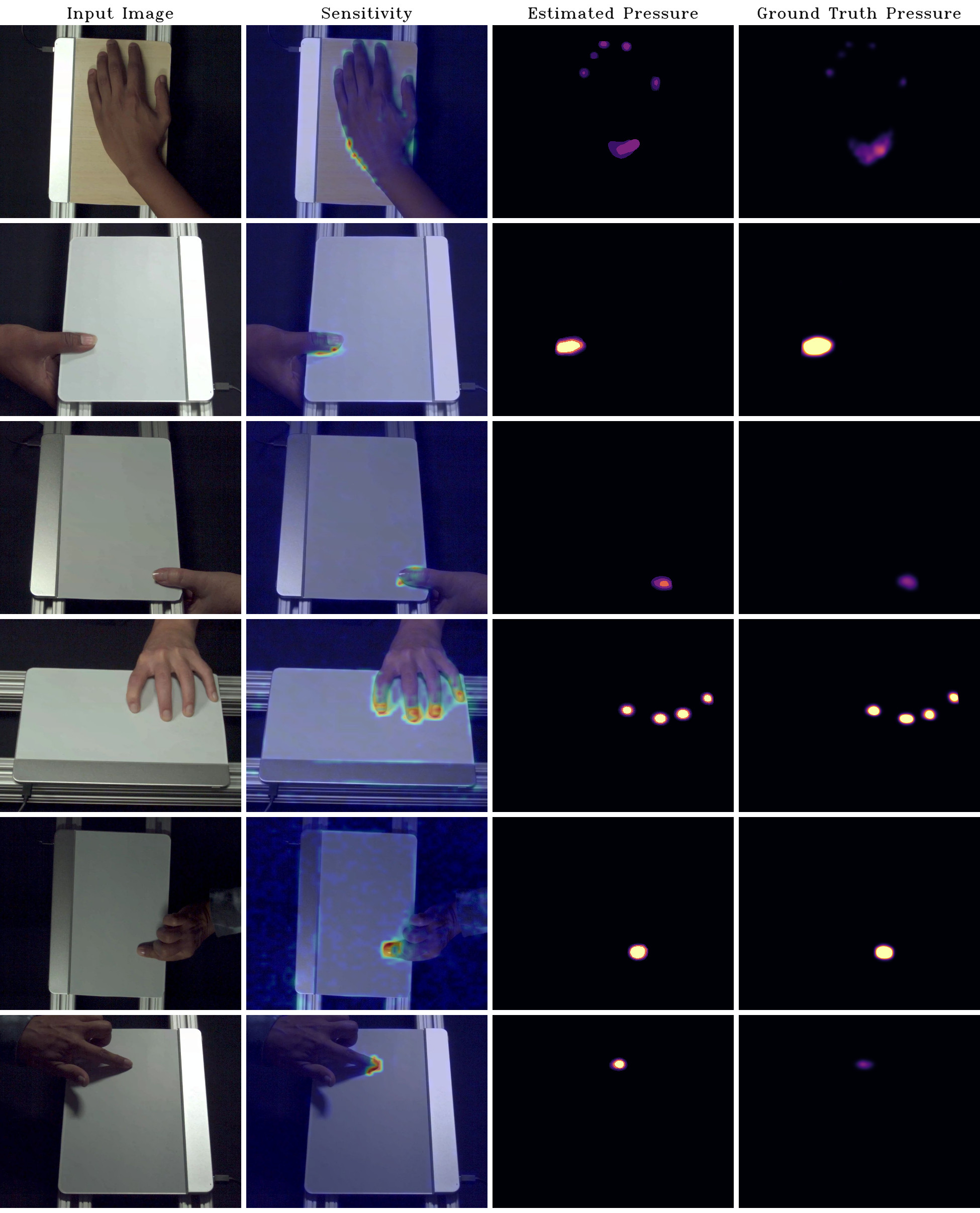}
\end{center}
\vspace{-8pt}
\caption{Results of PressureVisionNet selected \textit{randomly} from the test set.}
\label{fig:collage_1}
\end{figure*}

\begin{figure*}
\begin{center}
  \includegraphics[width=1.0\linewidth]{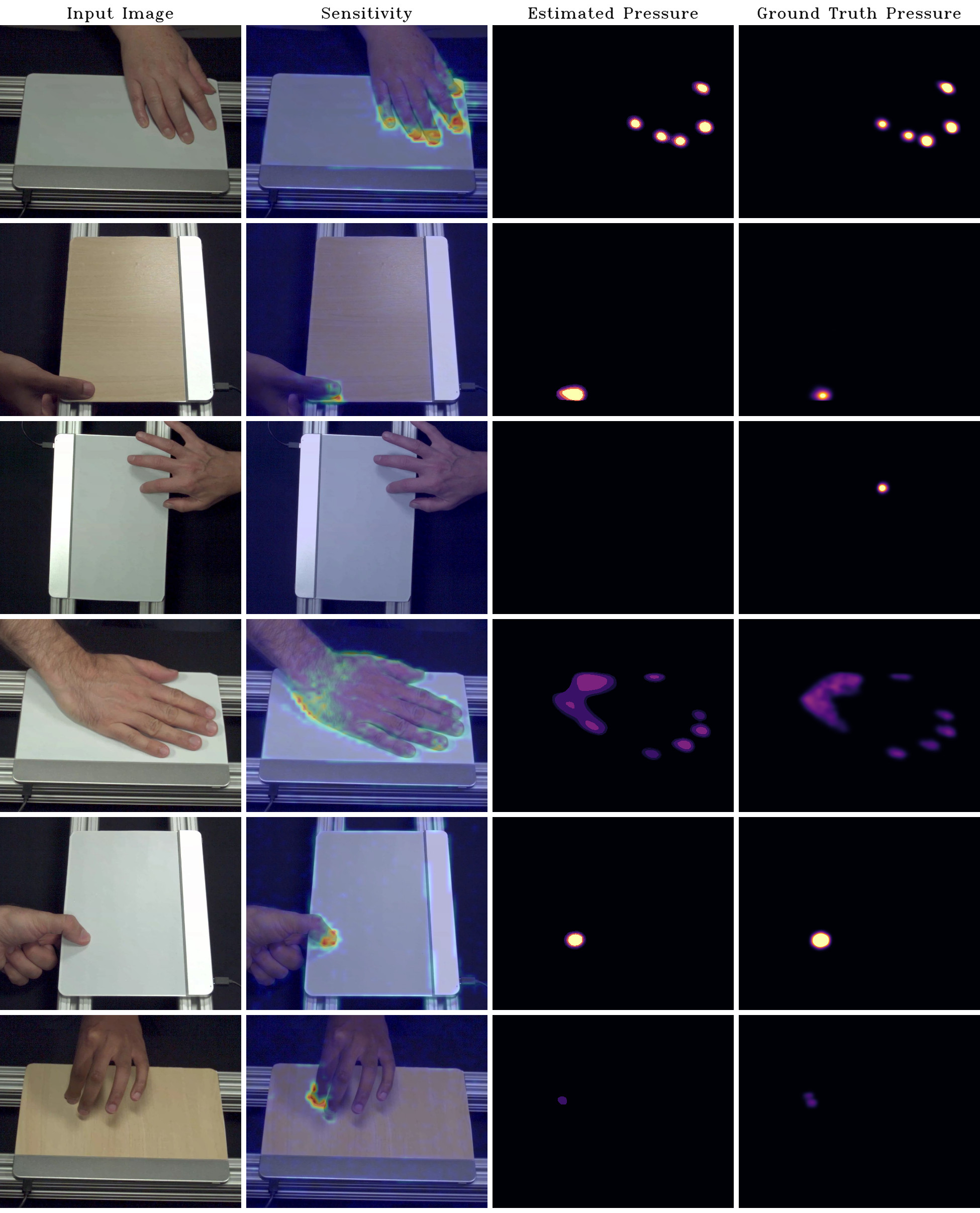}
\end{center}
\vspace{-8pt}
\caption{Results of PressureVisionNet selected \textit{randomly} from the test set.}
\label{fig:collage_2}
\end{figure*}

\begin{figure*}
\begin{center}
  \includegraphics[width=1.0\linewidth]{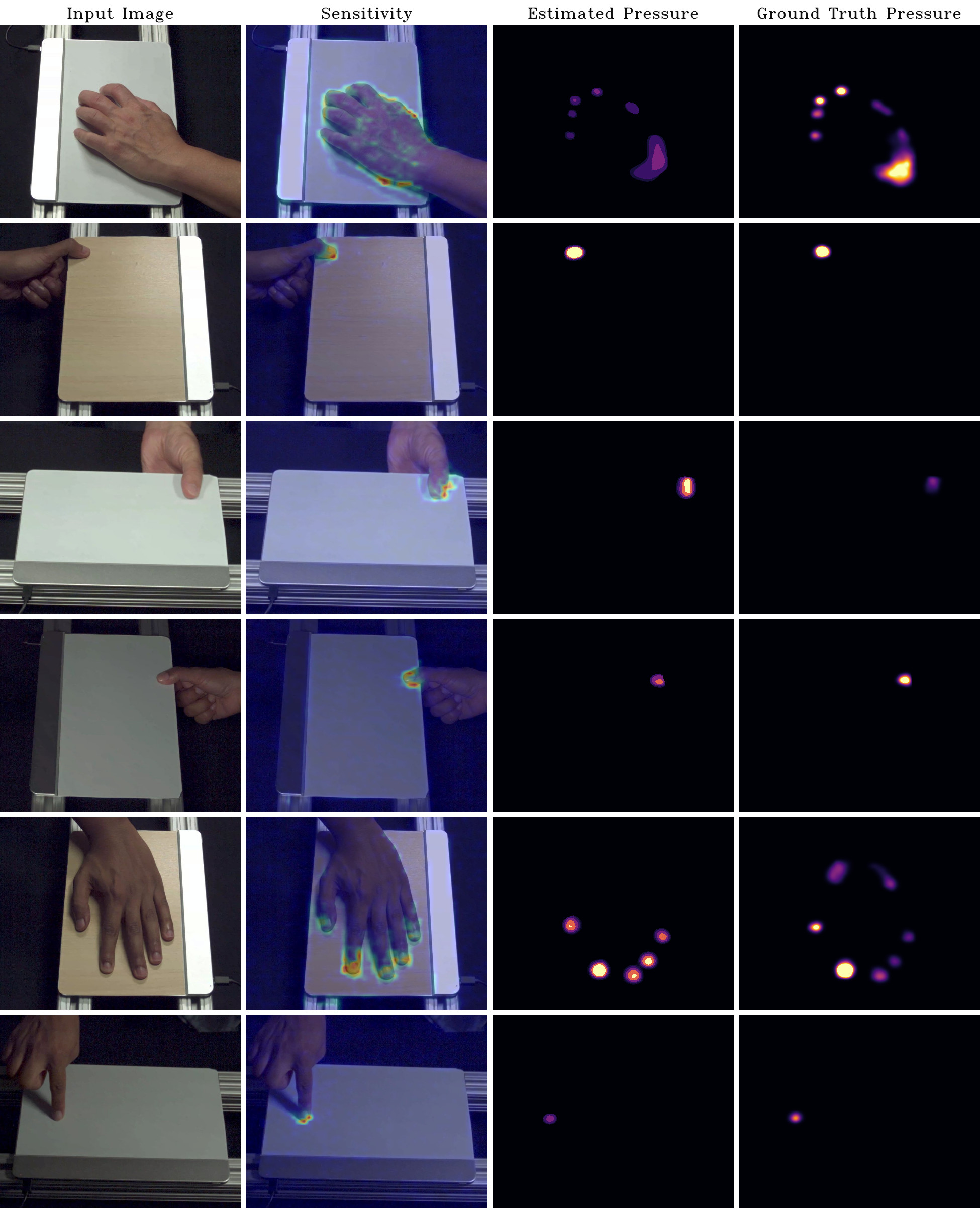}
\end{center}
\vspace{-8pt}
\caption{Results of PressureVisionNet selected \textit{randomly} from the test set.}
\label{fig:collage_3}
\end{figure*}

\begin{figure*}
\begin{center}
  \includegraphics[width=1.0\linewidth]{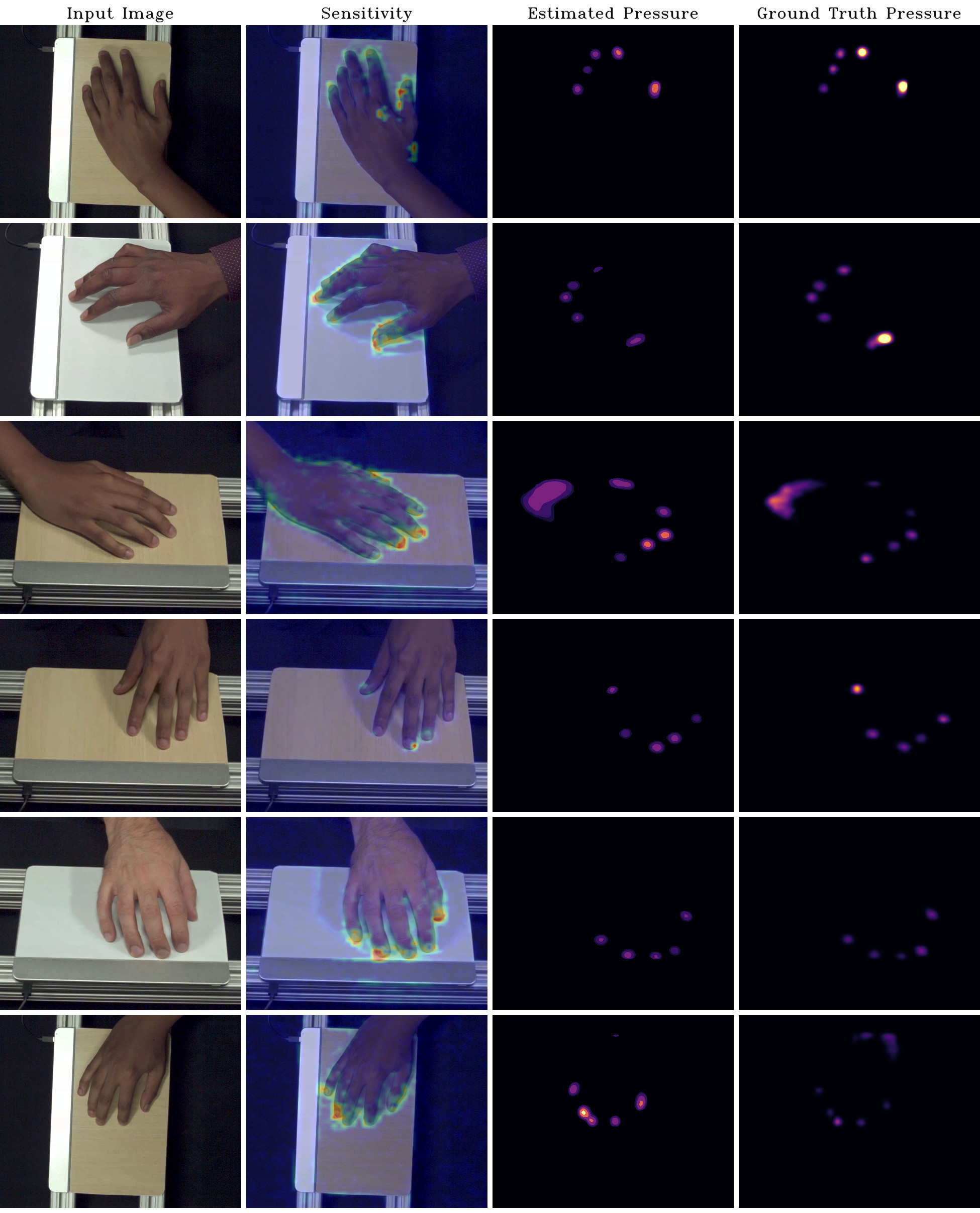}
\end{center}
\vspace{-8pt}
\caption{Results of PressureVisionNet selected \textit{randomly} from the test set.}
\label{fig:collage_4}
\end{figure*}

\begin{figure*}
\begin{center}
  \includegraphics[width=1.0\linewidth]{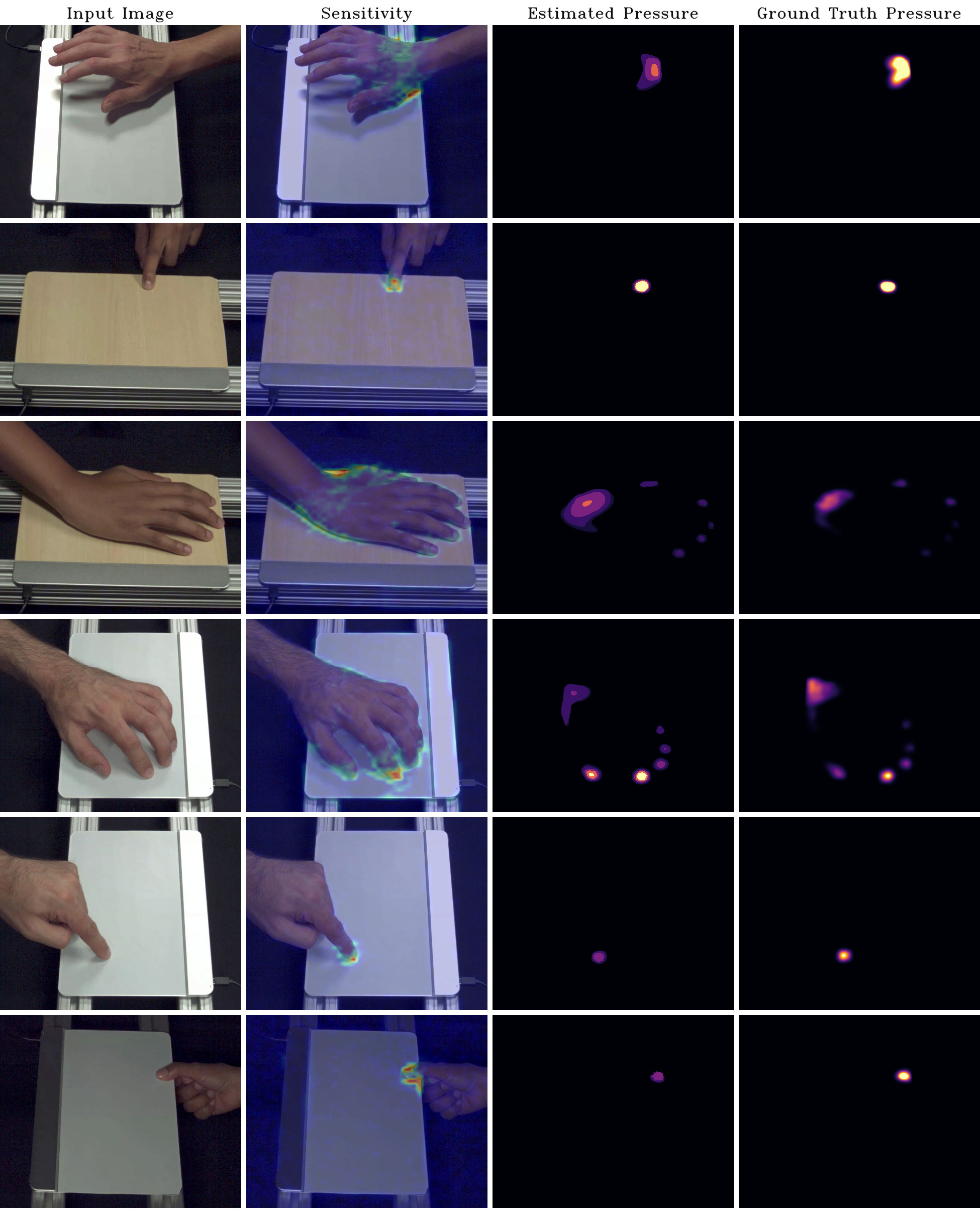}
\end{center}
\vspace{-8pt}
\caption{Results of PressureVisionNet selected \textit{randomly} from the test set.}
\label{fig:collage_5}
\end{figure*}

Further results from PressureVisionNet are included in Figures \ref{fig:collage_0}-\ref{fig:collage_5}. These results are \textit{randomly} selected from frames in the test set. As there a large number of frames where the hand is not present or is not near the object, frames with zero total ground truth pressure are omitted.

\subsection{Temporal PressureVisionNet}
Motion can be useful for machine perception of human activities.
To evaluate if motion provides salient cues for visual hand pressure estimation, we extend our base model to incorporate temporal information. The PressureVisionNet (PV-Net) structure is modified by concatenating the encoded features from multiple frames before the decoder module. The encoder networks have tied weights. The network was trained and tested on sequences of 4 frames, 0.2 sec apart. As expected, this temporal network outperforms our base model (Table \ref{tab:supp_temporal_results}). However, due to the ease of use and similar performance of the single-frame model, we focus on this for all experiments. Notably, the single-frame model may still leverage some motion cues by perceiving motion blur present in our RGB images.

\begin{table*}
\caption{A version of PressureVisionNet incorporating temporal information slightly outperforms our single-frame model.}
\centering
\begin{tabular}{c|c|c|c|c|c}
    \textbf{Method} & \textbf{Frames}& \textbf{Temporal Acc} & \textbf{Contact IoU} & \textbf{Vol. IoU} & \textbf{MAE}\\\hline
    PV-Net & 1 & 96.2\% & 55.8\% & 41.3\% & 39.9 Pa \\\hline
    PV-Net-Temporal & 4 & \textbf{96.8\%} & \textbf{56.6\%} & \textbf{43.1\%} & \textbf{39.4 Pa} \\\hline
\end{tabular}
\label{tab:supp_temporal_results}
\end{table*}

\subsection{Degraded Imagery} \label{sec:supp_degraded}
We also evaluated the performance of our network trained and tested with degraded images (Table \ref{tab:monochrome_results}). PressureVisionNet uses 480x384 images around the contact surface, captured in a controlled, well-lit setting. We show the performance of a model trained and tested with monochrome images. The drop in Volumetric IoU performance suggests that the model uses color when inferring pressure, but other information is sufficient for contact estimation. We also evaluated models trained and tested with lower resolution images. There was a modest drop in performance until the resolution went below 120x96 pixels. This suggests that our approach may be applicable to images captured in a less controlled setting in which hands occupy a smaller part of the image. 

\begin{table}
\caption{Performance decreases slowly as resolution decreases. Evaluating on monochrome images shows a reduction in Volumetric IoU.}
\centering
\begin{tabular}{c|c|c|c|c}
    \textbf{Method} & \textbf{Temporal Acc} & \textbf{Contact IoU} & \textbf{Volumetric IoU} & \textbf{MAE}\\\hline
    Zero Guesser & 53.7\% & 0.0\% & 0.0\% &  51.9 Pa \\\hline\hline
    RGB-15x12 & 78.9\% & 17.6\% & 11.7\% & 55.0 Pa \\\hline
    RGB-30x24 & 91.4\% & 38.5\% & 28.0\% & 49.6 Pa \\\hline
    RGB-60x48 & 94.9\% & 50.5\% & 36.2\% & 45.5 Pa \\\hline
    RGB-120x96 & 95.8\% & 53.6\% & 39.5\% & 42.8 Pa \\\hline
    RGB-240x192 & 95.9\% & 54.2\% & 39.3\% & 40.2 Pa \\\hline
    Mono-480x384 & 96.1\% & 53.2\% & 37.7\% & 40.9 Pa \\\hline\hline
    \begin{tabular}{@{}c@{}}PressureVisionNet\\RGB-480x384\end{tabular} & \textbf{96.2\%} & \textbf{55.8\%} & \textbf{41.3\%} & \textbf{39.9 Pa} \\\hline
\end{tabular}
\label{tab:monochrome_results}
\end{table}

\newpage

\begin{table*}
\caption{The list of actions executed for each hand during the collection of PressureVisionDB. For actions where the \textit{force level} is given, the participant is prompted to apply the specified amount of force. For actions where it is not specified, the participant may apply any force level.}
\label{tab:actions}
\centering
\begin{tabular}{l|c|c}
    \textbf{Action name} & \textbf{Repetitions} & \textbf{Force Level}\\\hline
    Calibration routine & 1 & - \\\hline
    Press index finger & 5 & Low \\\hline
    Press index finger & 5 & High \\\hline
    Press index finger & 5 & No Contact \\\hline
    Index finger, pull towards & 5 & - \\\hline
    Index finger, push left & 5 & - \\\hline
    Index finger, push away & 5 & - \\\hline
    Index finger, push right & 5 & - \\\hline
    Press palm & 5 & Low \\\hline
    Press palm & 5 & High \\\hline
    Press palm & 5 & No Contact \\\hline
    Press palm and fingers & 5 & Low \\\hline
    Press palm and fingers & 5 & High \\\hline
    Press palm and fingers & 5 & No Contact \\\hline
    Press fingers & 5 & Low \\\hline
    Press fingers & 5 & High \\\hline
    Press fingers & 5 & No Contact \\\hline
    Press finger one by one, flat & 3 & Low \\\hline
    Press finger one by one, flat & 3 & High \\\hline
    Press finger one by one, cupped & 3 & Low \\\hline
    Press finger one by one, cupped & 3 & High \\\hline
    Fingertips down, push away & 5 & - \\\hline
    Fingertips down, pull towards & 5 & - \\\hline
    Grasp edge, thumb down, uncurled fingers & 5 & - \\\hline
    Grasp edge, thumb up, uncurled fingers & 5 & - \\\hline
    Grasp edge, thumb down, curled fingers & 5 & - \\\hline
    Grasp edge, thumb up, curled fingers & 5 & - \\\hline
    Pinch, thumb down & 5 & Low \\\hline
    Pinch, thumb down & 5 & High \\\hline
    Pinch, thumb down & 5 & No contact \\\hline
    Pinch, thumb up & 5 & Low \\\hline
    Pinch, thumb up & 5 & High \\\hline
    Pinch, thumb up & 5 & No contact \\\hline
    Draw word & 5 & - \\\hline
    Pinch-zoom & 5 & - \\\hline
\end{tabular}
\end{table*}



\end{document}